\documentclass[10pt,letterpaper]{article}
\usepackage[top=0.85in,left=2.75in,footskip=0.75in]{geometry}

\usepackage{booktabs}
\usepackage{amsmath,amssymb}

\usepackage{changepage}

\usepackage[utf8x]{inputenc}

\usepackage{textcomp,marvosym}

\usepackage{cite}

\usepackage{nameref,hyperref}

\usepackage[right]{lineno}

\usepackage{microtype}
\DisableLigatures[f]{encoding = *, family = * }

\usepackage[table]{xcolor}

\usepackage{array}

\newcolumntype{+}{!{\vrule width 2pt}}

\newlength\savedwidth

\raggedright
\usepackage[aboveskip=1pt,labelfont=bf,labelsep=period,justification=raggedright,singlelinecheck=off]{caption}

\makeatletter
\renewcommand{\@biblabel}[1]{\quad#1.}
\makeatother

\date{}

\usepackage{lastpage,fancyhdr,graphicx}
\usepackage{epstopdf}
\fancyheadoffset[L]{2.25in}
\fancyfootoffset[L]{2.25in}

\begin{document}
\vspace*{0.2in}

\begin{flushleft}
{\Large
\textbf\newline{Predicting and Understanding Law-Making with Word Vectors and an Ensemble Model} 
}
\newline
\\
John J. Nay*\textsuperscript{1,2}
\\
\bigskip
\textbf{1} School of Engineering, Vanderbilt University, Nashville, TN, United States of America
\\
\textbf{2} Program on Law and Innovation, Vanderbilt Law School, Nashville, TN, United States of America
\bigskip

* john.j.nay@gmail.com

\end{flushleft}


\section*{Abstract}
Out of nearly 70,000 bills introduced in the U.S. Congress from 2001 to
2015, only 2,513 were enacted. We developed a machine learning approach
to forecasting the probability that any bill will become law. Starting
in 2001 with the 107th Congress, we trained models on data from
\emph{previous} Congresses, predicted all bills in the \emph{current}
Congress, and repeated until the 113th Congress served as the test. For
prediction we scored each sentence of a bill with a language model that
embeds legislative vocabulary into a high-dimensional, semantic-laden vector space. This
language representation enables our investigation into which words
increase the probability of enactment for any topic. To test the
relative importance of text and context, we compared the text model to a
context-only model that uses variables such as whether the bill's
sponsor is in the majority party. To test the effect of changes to bills
after their introduction on our ability to predict their final outcome,
we compared using the bill text and meta-data available at the time of
introduction with using the most recent data. At the time of
introduction context-only predictions outperform text-only, and with the
newest data text-only outperforms context-only. Combining text and
context always performs best. We conducted a global sensitivity analysis
on the combined model to determine important variables predicting
enactment.


\section{Introduction}\label{introduction}

The U.S. legislative branch creates laws that impact the lives of
hundreds of millions of citizens. For example, the Patient Protection and Affordable
Care Act (ACA) significantly affected the health care industry and
individuals' health insurance coverage. Bills often consist of hundreds
of pages of dense legal language. In fact, the ACA is more than 900
pages long. There are thousands of bills under consideration at any
given time and only about 4\% will become law. Furthermore, the number
of bills introduced is trending upward (see S1 Appendix), exacerbating the
problem of determining what text is relevant. Given the complexity,
length, and vast quantity of bills, a machine learning approach that
leverages bill text is well-suited to forecast bill success and identify
the important predictive variables. Despite rapid advancement of machine
learning methods, it's difficult to outperform naive forecasts of rare
events because of inherent variability in complex social processes \cite{martin_exploring_2016}
and because relationships learned from historical data can change
without warning and invalidate models applied to future circumstances.

Due to the complexity of law-making and the aleatory uncertainty in the
underlying social systems, we predict enactment probabilistically. It's
important to make \emph{probabilistic} predictions for high consequence
events because even small changes in probabilities for events with
extreme implications can have large expected values. For instance, the 2009 stimulus bill cost \$831 billion so even a 0.1 change in the predicted probability of this bill corresponds to a \$83.1 billion dollar change in the expected value (the probability of an event multiplied by its consequences). Probabilities
provide much more information than a simple ``enact'' or ``not enact''
prediction. Model performance metrics that don't use probabilities, such
as accuracy, are not suitable measures of rare event predictive ability.
For instance, a blunt ``never enact'' model has a seemingly impressive
96\% accuracy rate on this data but incorrectly classifies all the
enacted bills with incalculable effects on society.

Forecasting model performance should be estimated using multiple metrics
on large amounts of test data measured \emph{after} the data that was
used to train the model. We trained models on Congresses prior to the
Congress predicted, which simulated real-time deployment across 14 years
and 68,863 bills. Starting with the 107th Congress (2001--2003), models
were sequentially trained on data from \emph{previous} Congresses and
tested on all bills in the \emph{current} Congress. This was repeated seven times
until the most recently completed Congress -- the 113th (2013--2015) --
served as the test. To estimate performance, we compared a baseline
model to our models across three performance measures that leverage
predicted probabilities.

Although previous research found that bill text was useful for
predicting whether bills will survive committee \cite{yano_textual_2012} and for predicting
roll call votes \cite{wang_joint_2010, gerrish_predicting_2011}, these authors tested their models on much less
data than we do and predicted more frequently observed events: getting
out of committee is more common than being enacted and bills up for vote
are a small subset of all bills introduced. It's not clear whether
utilizing text models trained on previous Congresses will improve
predictions of enactment of bills introduced in future Congresses beyond
the predictive power of sponsorship, committee and other non-textual
data. Text is noisy and completely different topics can be found within
the same bill \cite{wilkerson_tracing_2015}. However, we hypothesized that there are unique
semantic and syntactic signatures that consistently differentiate successful bills.
Our second hypothesis was concerned with the changes to bills over their
lives. Some bills, e.g.~the ACA, are only a few pages when introduced
but are hundreds of pages when enacted. However, 87\% of bill texts
don't change after being introduced because they don't progress further
in the law-making process. We hypothesized that using the most recently
available version of bill text and metadata would lead to stronger
predictive performance for text and context models. To test these
hypotheses, we designed an experiment across two primary dimensions:
\emph{data type} (text-only, text and context, or context-only) and
\emph{time} (using oldest or newest bill data).

Analyzing a model that makes successful ex ante predictions can be more
informative than ex-post interpretations of socio-political events
(outside experiment-like settings) due to the over-fitting that plagues
most modeling of observational data \cite{katz_predicting_2014}. However, because highly
predictive models are often designed with only predictive power in mind,
they rarely provide clear insights into relationships between predictor
variables and the predicted outcome. When estimates of these
relationships are provided for non-linear models, they are almost always
measures of only magnitudes of the effects of predictor variables and
not also the directions of the effects. Our work is not limited to raw
predictive power. We estimate the \emph{direction and magnitude} of the
effect of each predictor variable in the model on the predicted
probability of enactment. Furthermore, the text model reveals which
words are more associated with enactment success.

\section{Methods and Data}

\subsection{Model Training}\label{model-training}

\subsubsection{Word Vectors and Inversion of Language
Models}\label{word-vectors-and-inversion-of-language-models}

Continuous-space vector representations of words can capture subtle
semantics across the dimensions of the vector \cite{nay_gov_2016}. To learn these
representations, a neural network model predicts a target word with the
mean of the representations of the surrounding words (e.g.~vectors for
the two words on either side of the target word in Fig. 1A.). The
prediction errors are then back-propagated through the network to update
the representations in the direction of higher probability of observing
the target word \cite{mikolov_distributed_2013, bengio_neural_2003}. After randomly initializing representations and
iterating this process over many word pairings, words with similar
meanings are eventually located in similar locations in vector space as
a by-product of the prediction task, which is called word2vec \cite{mikolov_distributed_2013}.

\begin{figure}[h]
\includegraphics[width=2.1in]{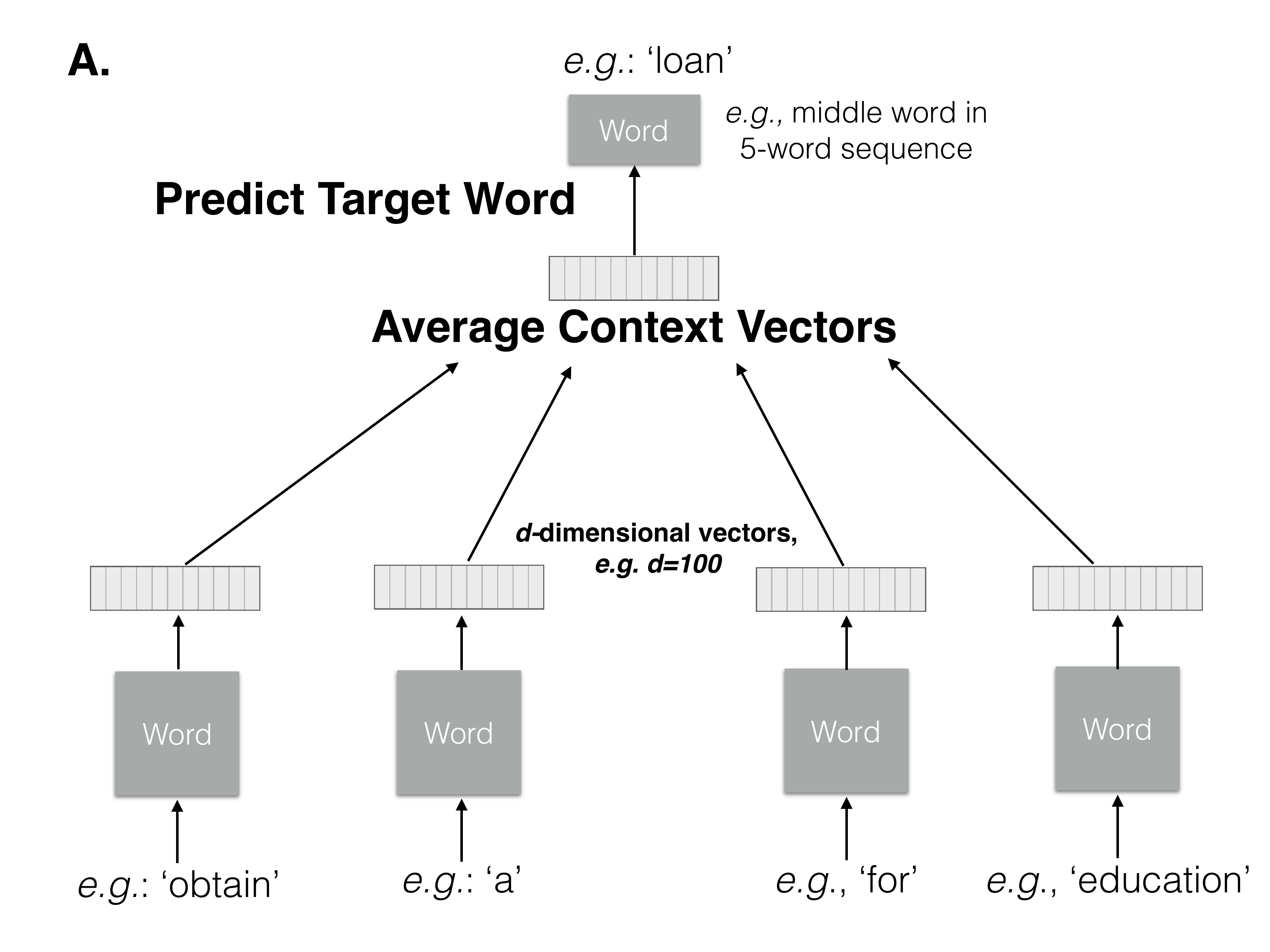}
\includegraphics[width=2.1in]{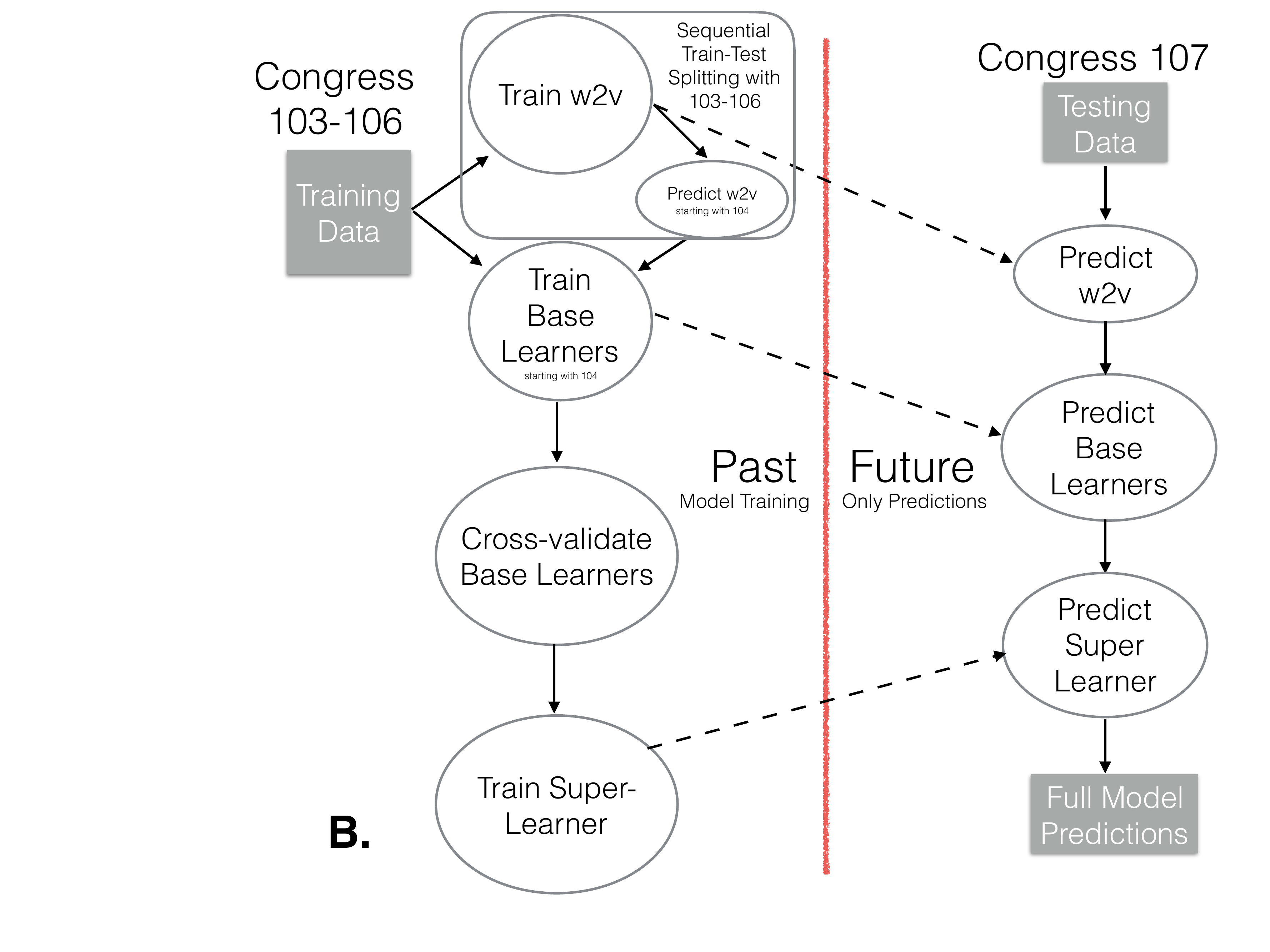}
\caption{ \textbf{A.} The neural network-based training algorithm
used to obtain word vectors \cite{rehurek_software_2010}. Parameters are updated with stochastic gradient descent
and we use a binary Huffman tree to implement efficient softmax
prediction of words. See S1 Appendix for description of hyper-parameters. \textbf{B.} Model
training and testing. This process is completed and then we
advance one Congress.}
\end{figure}

The only pre-processing we applied to text was removal of HTML, carriage
returns, and whitespace, and conversion to lower-case. Then inversion of
distributed language models was used for classification as described in
\cite{taddy_document_2015}. Distributed language models -- mappings from words to
\(\mathbf{R}^d\) obtained by leveraging word co-occurrences -- were
separately fit to the sub-corpora of successful and failed bills by
applying word2vec. Each sentence of a testing bill was scored with each
trained language model and Bayes' rule was applied to these scores and
prior probabilities for bill enactment to obtain posterior
probabilities. The proportions of bills enacted in the same chamber as
the predicted bill in all \emph{previous} Congresses were used as the
priors. The probabilities of enactment were then averaged across all
sentences in a bill to assign an overall probability.

\subsubsection{Tree-based Models}\label{tree-based-models}

Trees are decision rules that divide predictor variable space into
regions by choosing variables and their threshold values on which to
make binary splits \cite{breiman_classification_1984}. A tree model can learn interactions between
predictors, unlike linear models where interactions must be manually
specified, and is generally robust to the inclusion of variables
unrelated to the outcome. A \emph{gradient boosted machine} (GBM)
improves an ensemble of weaker base models, often trees, by sequentially
adjusting the training data based on the residuals of the preceding
models \cite{friedman_greedy_2001}. A \emph{random forest} randomly samples observations from
training data and grows a tree on each sample, forcing each tree to
consider randomly selected sets of predictor variables at each split to
reduce correlation between trees \cite{breiman_random_2001}. GBMs and random forests can both
learn non-linear functions but have different strengths: in general,
random forests are more robust to outliers while GBMs can more
effectively learn complex functions. A regularized logistic
regression (elastic-net) with hyper-parameters (\(\alpha\) = 0.5
and \(\lambda\) = 1e-05) is also estimated to gain a complementary
linear perspective \cite{www.h2o.ai_h2o.ai_2016}.

Using the predictions from the inversion of the word vector language model (as described in Section 2.1.1.) as features allows the training process to learn interactions between contextual variables and textual
probabilities. Additionally, the sensitivity analysis can then estimate
the impact of text predictions on enactment probabilities along with the
contextual predictors, controlling for the effect of the probability of
the bill text when estimating non-textual effects.

\subsubsection{Ensemble Stacking}\label{ensemble-stacking}

Random forests and GBMs combine weak learners to create a strong
learner. Stacking combines strong learners to create a stronger learner.
A cross-validation stacking process on the training data is used to
learn a combination of the three base models to form a meta-predictor
\cite{breiman_stacked_1996, van_super_2007}. Out-of-fold cross-validation predictions are made on the training data with the three base learners described above in Section 2.1.2 (the gradient boosted machine model, the random forest model, and the regularized logistic regression model).
These predictions and the outcome vector are used to train the
meta-learner, a regularized logistic regression with non-negative
weights. Weights are forced to be non-negative because we assume all
predictors should positively contribute. This entire learning process is
conducted on data from prior Congresses. The model is applied to test
data by making predictions with base learners and feeding those into the
meta-learner (Fig. 1B.).

\subsection{Model Performance}\label{model-performance}

We use the two most frequently applied binary classification probability
scoring functions: the log score and the brier score (see S1 Appendix). For both,
if a model assigns high probability to a failed bill it's penalized more
than if it was less confident and if a model assigns high probability to
an enacted bill it is rewarded more than if it wasn't confident. A
receiver operating characteristic curve (ROC) is built from points that
correspond to the true positive rate at varying false positive rate
thresholds with the model's predictions sorted by the probability of the
positive class (enacted bill) \cite{altman_diagnostic_1994}. Starting at the origin of the space
of true positive rate against false positive rate, the prediction's
impact on the rates results in a curve tracing vertically for a correct
prediction and horizontally for an incorrect prediction. A perfect area
under the ROC curve (AUC) is 1.0 and the worst is 0.5. AUC rewards models for being discriminative throughout the range of probabilities and is more appropriate than accuracy for imbalanced
datasets.

\subsection{Analysis}\label{analysis-1}

\subsubsection{Text Model Similarity
Analysis}\label{text-model-similarity-analysis}

We train language models with word2vec for enacted House bills, failed
House bills, enacted Senate bills, and failed Senate bills and then
investigate the most similar words within each of these four models to
word vector combinations representing topics of interest. That is, for
each of the four models, return a list of most similar words:
\(\arg\max_{v* \in V_{1:N}}{cos(v*, \frac{1}{W} \sum_{i=1}^{W} w_i \times s_i)}\),
where \(w_i\) is one of \(W\) word vectors of interest, \(V_{1:N}\) are
the \(N\) most frequent words in the vocabulary of \(M\) words (rare
words are retained to train the model, but \(N\) is set to less than
\(M\) to exclude rare words during model analysis) excluding words
corresponding to the \(W\) query vectors, \(s_i\) is \emph{1} or
\emph{-1} for whether we are positively or negatively weighting \(w_i\),
and
\(cos(a, b) = \frac{\sum_{i=1}^{d} a_i \times b_i}{\sqrt{\sum_{i=1}^{d} a_i^2} \sqrt{\sum_{i=1}^{d} b_i^2}}\).
For ease of comparison across enacted and failed categories we also
remove words the two have in common.

\subsubsection{Full Model Sensitivity
Analysis}\label{full-model-sensitivity-analysis}

We conduct a sensitivity analysis on our model of the legislative
system by varying inputs to the model and measuring the effect on the
output. If input values are varied one at a time, while keeping the
others at ``default values,'' sensitivities are conditional on the
chosen default values \cite{saltelli_how_2010}. There are no sensible default values for the
predictor variables. Instead of using default values of variables, we use empirically observed predictor variables and we predict the enactment of all the bills from Congresses 104-112. These predictions are a
vector of predicted probabilities. The empirical predictors variable data and these associated predicted probabilities create a
sufficiently large yet realistic set of observations for a global sensitivity analysis to determine the modeled effects of the predictors variables on the probability of enactment.

Next, we expand the factor variables out so each level is represented in
the design matrix as a binary indicator variable. This allows us to
estimate the effect of each level of a factor, e.g.~the 39 subject
categories. We add interaction terms between the Chamber and bill
characteristics, e.g.~whether the bill originated in the Senate and the
number of characters, to estimate these interaction effects potentially
automatically learned by the tree models. Finally, we estimate the
relationship between the resulting matrix of input values and the vector
of predicted probability outputs with a partial rank correlation
coefficient (PRCC) analysis, which estimates the correlation between an
input variable and the predicted probability of bill enactment,
discounting the effects of the other inputs and allowing for potentially
non-linear relationships by rank-transforming the data before model
estimation \cite{pujol_sensitivity:_2014, saltelli_sensitivity_2009}. Partial correlation controls for the other
predictor variables, \(Z\), by computing the correlation between the
\emph{residuals} of regressing the predictor of interest, \(x\), on
\(Z\) and the \emph{residuals} of regressing the outcome (predicted
probability of enactment) on \(Z\). The PRCC analysis is bootstrapped
1,000 times to obtain 95\% confidence intervals.

\subsection{Data}\label{data}

We include all House and Senate bills and exclude simple, joint, and
concurrent resolutions because simple and concurrent resolutions do not
have the force of law and joint resolutions are very rare. We downloaded
all bill data (from the 103rd Congress through the 113th Congress) other than committee membership from
govtrack.us/developers/data, which is created by scraping THOMAS.gov. We
downloaded committee membership data from
web.mit.edu/17.251/www/data\_page.html \cite{stewart_iii_congressional_2011, stewart_iii_congressional_2016}.

There is often more than one version of the full text for each bill. In
order to create a forecasting problem that predicts enactment as soon as
possible, the earliest dated full text is used, which is, for more than
99\% of the bills in the testing data, the text as it was introduced. To
understand how much predictive power newer versions add, we collect the
most recent version of each bill, which is, for 87\% of the bills in the
testing data, the version as introduced. Bills can change dramatically between the time of their introduction and the time of the last action taken on them. H.R. 3590 in the 111th Congress, was a short bill on housing tax changes for service members when it was introduced, and shortly before it was enacted it was the 906-page Affordable Care Act. H.R. 34 in the 114th Congress was originally introduced as the Tsunami Warning, Education, and Research Act and was about 30 pages long. Shortly before it was enacted, H.R. 34 was the 312-page 21st Century Cures Act.

The full text of all introduced bills is only available starting with
the 103rd Congress (1993--1995) and therefore this is the first Congress
used to train language models. The 104th Congress is the first used to
train the base models of the ensemble because they require the language
model predictions and the language models need the 103rd for training.
The 107th Congress (2001--2003) is the first to serve as a testing
Congress because the full model needs multiple Congresses worth of data
for training. We used the list of predictor variables from \cite{yano_textual_2012} as a
starting point for designing our feature set.

The following variables capture characteristics of a bill's sponsor and
committee(s):

\begin{itemize}
\item
  \emph{region}: region corresponding to state the sponsor represents (5
  levels).
\item
  \emph{sponsorPartyProp}: proportion of chamber in sponsor's party
  (min: 0, median: 0.51, max: 0.59).
\item
  \emph{sponsorTerms}: number of terms sponsor has served in Congress
  (only up to Congress being predicted to ensure model is only using
  data that would have been available at that time, min: 1, median: 6,
  max: 30).
\item
  \emph{committeeSeniority}: mean length of time sponsor has been on the
  committees the bill is assigned to (min: 0, median: 0, max: 51). If
  not on committee, assigned 0.
\item
  \emph{committeePosition}: out of any leadership position of sponsor on
  any committee bill is assigned to, lowest number on the ``leadership
  codes'' list in S1 Appendix (11 levels, e.g.~Chairman).
\item
  \emph{NotMajOnCom}: binary for whether sponsor is (\emph{i})
  \emph{not} in majority party and (\emph{ii}) on first listed committee
  bill is assigned to.
\item
  \emph{MajOnCom}: binary for whether sponsor is (\emph{i}) in majority
  party and (\emph{ii}) on first listed committee bill is assigned to.
\item
  \emph{numCosponsors}: number of co-sponsors (for oldest - min: 0,
  median: 2, max: 378; for newest - min: 1, median: 6, max: 432).
\end{itemize}

The following variables capture political and temporal context of bills:

\begin{itemize}
\item
  \emph{session}: Session (first or second) of Congress that corresponds
  to full text date, almost always the date bill was introduced for
  oldest data (for oldest - proportion in first session: 0.64; for
  newest - proportion in first session: 0.6).
\item
  \emph{house}: binary for whether it's a House bill.
\item
  \emph{month}: month bill is introduced.
\end{itemize}

The following variables capture aspects of bill content and
characteristics:

\begin{itemize}
\item
  \emph{subjectsTopTerm}: official top subject term (36 levels).
\item
  \emph{textLength}: number of characters in full text (for oldest -
  min: 119, median: 5,340, max: 2,668,424; for newest - min: 113,
  median: 5,454, max: 3,375,468).
\end{itemize}

\section{Results}\label{results}

\subsection{Prediction Experiments}\label{prediction-experiments}

Five models are compared across the two time conditions. \emph{w2v} is
the scoring of full bill text with an inversion of word2vec-learned
language representations \cite{taddy_document_2015}. We take this approach to textual
prediction because it provides the capacity to conduct a semantic
similarity text analysis across enacted and failed categories and can
predict which sentences of a bill contribute most to enactment.
\emph{w2vTitle} is title-only scoring with the same method. \emph{GLM}
is a regularized non-negative generalized linear model (GLM)
meta-learner over an ensemble of a regularized GLM, a gradient boosted
machine and a random forest, which each use only the contextual
variables (see Data section). \emph{w2vGLM} is the same as \emph{GLM}
except the \emph{w2v} and \emph{w2vTitle} predictions are added as two
more predictor variables for the three base learners. These are compared
to a baseline, \emph{null}, that uses the proportion of bills enacted in
the same chamber as the predicted bill across all previous Congresses as
the predicted probability. For instance, the proportion of bills enacted
in the Senate from the 103rd to the 110th Congress was 0.04 and so this
is the \emph{null} predicted probability of enactment of a Senate bill
in the 111th Congress. It's important to use Chamber-specific rates to
improve \emph{null} performance because bills originating in the House
have a higher enactment rate.

Using only text outperforms using only context on two of three
performance measures (AUC and Brier) for the newest data, while using
only context outperforms only text on three measures for the oldest data
(Fig. 2). Using text and context together, \emph{w2vGLM}, outperforms
all competitors on all measures for newest and oldest data (Table 1).
When predicting enactment with the newest bill text and the updated
number of cosponsors, text length and session, both models improved but
\emph{w2vGLM} and \emph{w2v} improved dramatically. \emph{w2vGLM} has
the highest AUC, \emph{w2v} has the second highest for predictions with
new data and \emph{GLM} has the second highest for predictions with old
data.

\begin{table}[]
\centering
\caption{ Model performance comparison (n=68,863). Lower mean brier score (MeanBrier)
and mean log loss (MeanLogLoss) is better and higher AUC is better. \emph{w2vTitle} has
infinite log loss due to making predictions with 0 and 1 probabilities.
Two-sample t-tests with alternative hypotheses that \emph{w2vGLM}
outperforms its closest competitors are significant with p-values of
3.02e-53 (log loss newest data), 3.65e-26 (brier loss newest data),
9.73e-04 (log loss oldest data) and 5.36e-03 (brier loss oldest data).}
\begin{tabular}{@{}llll@{}}
\toprule
Model       & AUC  & MeanBrier & MeanLogLoss \\ \midrule
w2vGLM      & 0.96 & 0.021     & 0.083       \\
w2v         & 0.93 & 0.027     & 0.127       \\
GLM         & 0.87 & 0.028     & 0.118       \\
w2vTitle    & 0.81 & 0.049     & Inf         \\
Null        & 0.58 & 0.035     & 0.157       \\
w2vGLMOld   & 0.85 & 0.029     & 0.122       \\
w2vOld      & 0.76 & 0.035     & 0.154       \\
GLMOld      & 0.83 & 0.031     & 0.131       \\
w2vTitleOld & 0.8  & 0.047     & Inf         \\ \bottomrule
\end{tabular}
\end{table}

\begin{figure}
\includegraphics{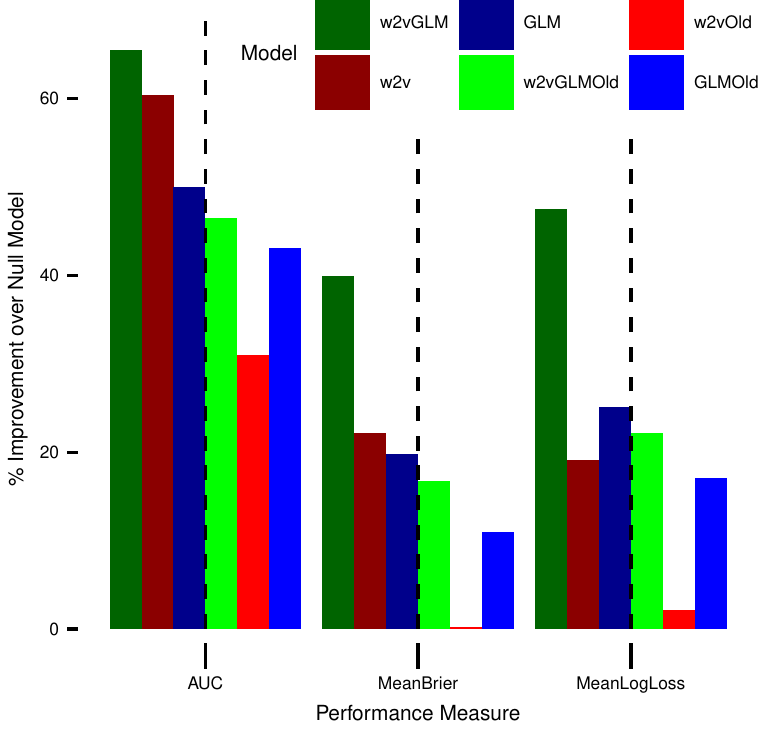}
\caption{Percent improvement of each model over \emph{null} (n=68,863).
Dashed lines separate newest and oldest data within each measure.
Because \emph{w2vTitle} has infinite log loss it's not included in the
figure.}
\end{figure}

Predicted probabilities of \emph{w2vGLM} range from 0.01 to 0.99. In
fact, the majority of the predicted probabilities are near 0 and 1 (S1 Appendix).
This is impressive given that it still maintains overall high
performance on log and brier scoring, which significantly penalize
models for high probability predictions on the wrong side of 0.5. The
central tendencies of the predicted probabilities (mean = 0.05, and
median = 0.01) are close to the observed frequency of bill enactment,
0.04. The median of the predicted probabilities where the true outcome
was not enact (0.01) is much lower than the median of the predicted
probabilities where the true outcome was enact (0.71) (Fig. 3A.). The
\emph{w2v} predicted probabilities (Fig. 3B.) demonstrate that with just
the text of the bills, the model can make probabilistic predictions that
discriminate between enacted and failed bills, providing credibility to
our textual semantic similarity analysis. In contrast, the title-only
(Fig. 3C.) and context-only (Fig. 3D.) models poorly discriminate.

\begin{figure}
\includegraphics[width=1.1in]{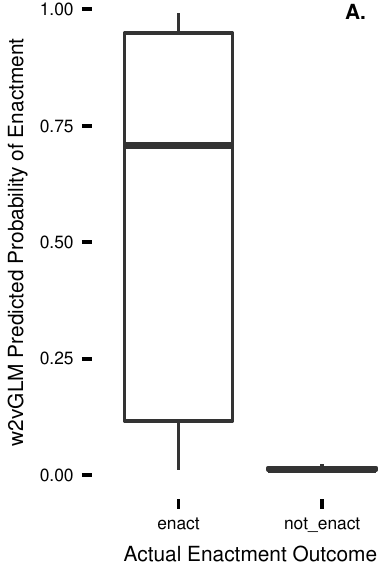}
\includegraphics[width=1.1in]{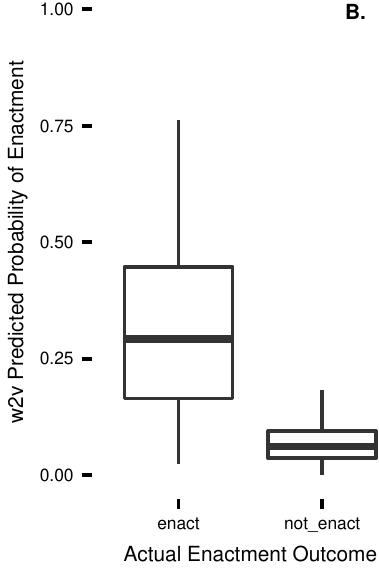}
\includegraphics[width=1.1in]{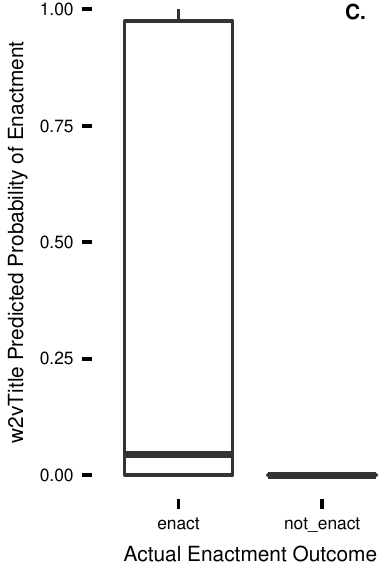}
\includegraphics[width=1.1in]{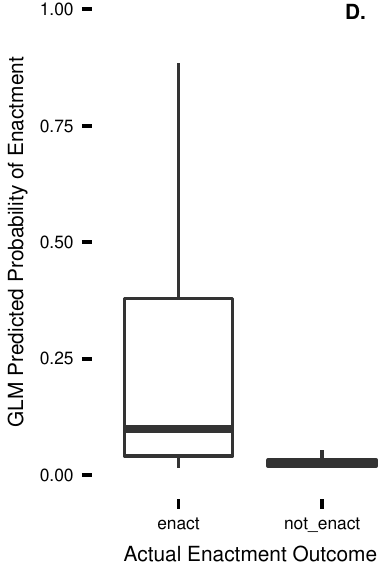}
\caption{Boxplots (n=68,863) of predicted probabilities of enacted and
failed bills for \emph{w2vGLM} (\textbf{A.}), \emph{w2v} (\textbf{B.}),
\emph{w2vTitle} (\textbf{C.}), \emph{GLM} (\textbf{D.}). The boxes are
the inter-quartile ranges (IQRs) of the predicted probabilities, the
bold line is the median, the whiskers extend from ends of IQR to
\(+/- 1.5 * IQR\).}
\end{figure}

We conduct an error analysis (see S1 Appendix) and find that, across all bill
subjects, \emph{w2vGLM} and \emph{w2v} both have the highest log loss on
two categories of low economic importance: Commemorations; and Arts,
Culture, Religion. As a final investigation of model performance, we
explore the \emph{w2vGLM} predictions for the two most significant
bills in the past century: the ACA and the American Recovery and
Reinvestment Act of 2009 (ARRA). The density of the predicted
probabilities for all bills is peaked around 0.01 (S1 Appendix) and the predicted
probabilities for the ACA and ARRA were \(>0.5\) (Table 2). None of the
296 other bills with ``Patient Protection and Affordable Care Act'' in
the title were enacted. These bills (all official titles listed in S1 Appendix)
attempted to amend or repeal the ACA, which could have had significant
economic effects. In 2012, the Congressional Budget Office estimated
that H.R.~6079, the Repeal of Obamacare Act, would cause a \$109 billion net increase
in federal deficits \cite{congressional_budget_office_letter_2012}. \emph{w2vGLM}'s predicted
probabilities for these failed attempts are much more useful than
\emph{null}'s (Table 2). 96\% of the (unsuccessful) bills to repeal and
amend the ACA have \emph{lower} predicted probabilities of enactment
from \emph{w2vGLM} than from \emph{null}.

\begin{table}[]
\centering
\caption{Predicted probabilities of enactment for key bills.
Probabilities increased between old and new forecasts for the two
enacted bills, and the mean of the probabilities for the failed bills
decreased.}
\begin{tabular}{@{}llll@{}}
\toprule
ShortTitle & ForecastNew & ForecastOld & BaselineForecast \\ \midrule
ACA & 0.6 & 0.23 & 0.05 \\
Failed Amend Repeal & 0.02 & 0.03 & 0.05 \\
ARRA & 0.55 & 0.52 & 0.05 \\ \bottomrule
\end{tabular}
\end{table}

\subsection{Analysis}\label{analysis}

Now that we have a model validated on thousands of predictions, we
analyze it to better understand law-making. With our language models, we create
``synthetic summaries'' of hypothetical bills by providing a set of
words that capture any topic of interest. Comparing these synthetic
summaries across chamber and across Enacted and Failed categories
uncovers textual patterns of how bill content is associated with
enactment. The title summaries are derived from investigating
similarities within \emph{w2vTitle} and the body summaries are derived
from similarities within \emph{w2v}. Distributed representations of the
words in the bills capture their meaning in a way that allows
semantically similar words to be discovered. Although bills may not have
been devoted to the topic of interest within any of the four training
data sub-corpora, these synthetic summaries can still yield useful
results because the queried words have been embedded within the
semantically structured vector space along with all vocabulary in the
training bills. This is important for a topic, such as climate change,
with little or no relevant enacted legislation.

To demonstrate the power of our approach, we investigated the words that
best summarize ``climate change emissions'', ``health insurance
poverty'', and ``technology patent'' topics for Enacted and Failed bills
in both the House and Senate (Fig. 4). ``Impacts,'' ``impact,'' and
``effects'' are in House Enacted while ``warming,'' ``global,'' and
``temperature'' are in House Failed, suggesting that, for the House
climate change topic, highlighting potential future impacts is
associated with enactment while emphasizing increasing global
temperatures is associated with failure. In both chambers,
``efficiencies'' is in Enacted and ``variability'' is in Failed. In the
Senate, ``anthropogenic'' (human-induced) and ``sequestration''
(removing greenhouse gases) are in Failed. For the health insurance
poverty topic, ``medicaid'' and ``reinsurance'' are in both House and
Senate Failed. The Senate has words related to more specific health
topics, e.g. ``immunization'' for Failed and ``psychiatric'' for
Enacted. For the patent topic, both chambers have a word related to
water (``fish'' and ``marine'') in the Failed Titles and ``geospatial''
in the Failed Bodies. Given recent legal developments regarding
patenting software, it's notable that ``software'' and ``computational''
are in Failed for the House and Senate, respectively.

\begin{figure}
\includegraphics[width=5.3in]{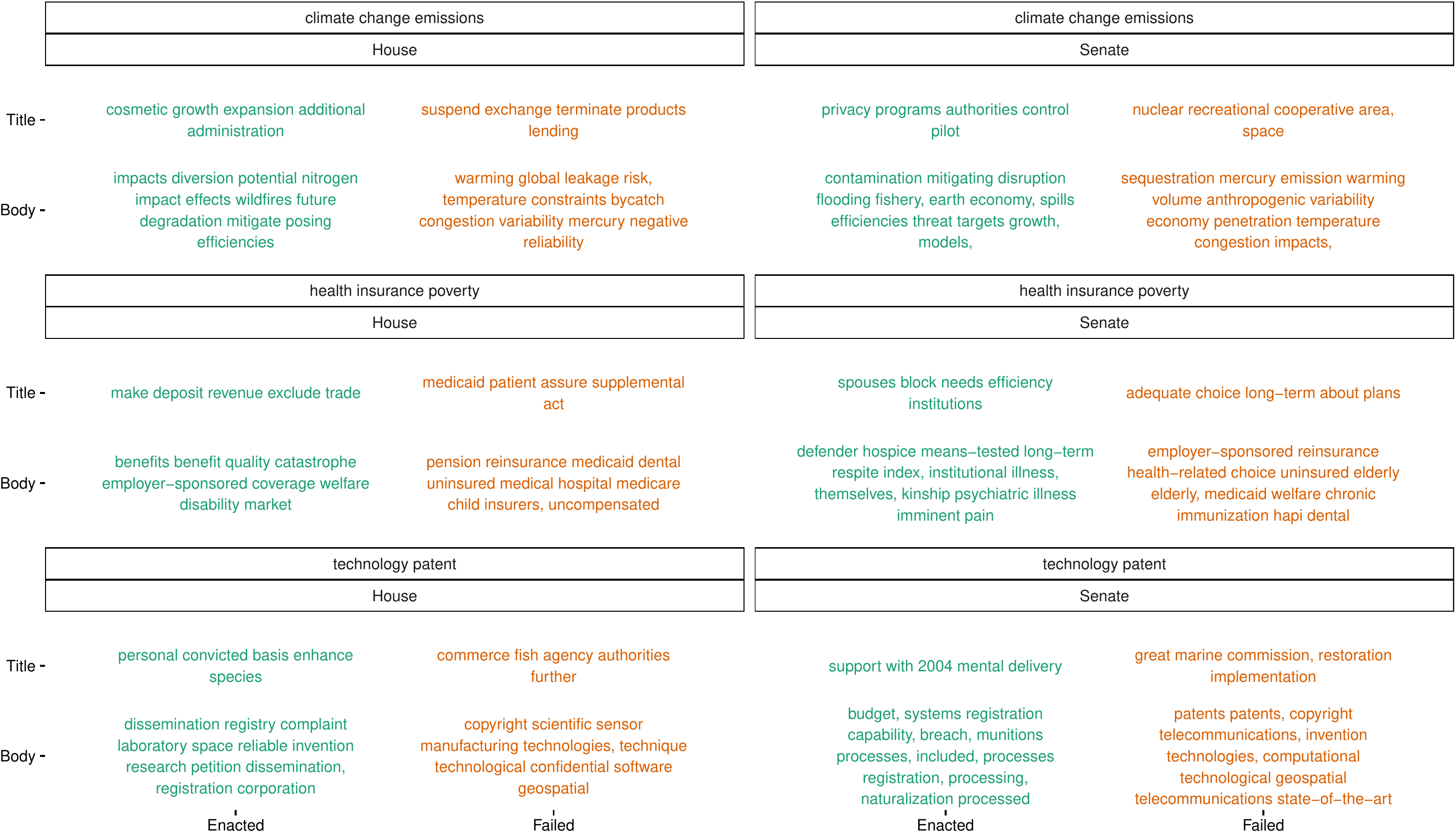}
\caption{Synthetic summary bills for three topics across Enacted
and Failed and House and Senate categories.}
\end{figure}

Our language model provides sentence-level predictions for an overall
bill and thus predicts what sections of a bill may be the most important
for increasing or decreasing the probability of enactment. Fig. 5
compares patterns of predicted sentence probabilities as they evolve
from the beginning to the end of bills across four categories: enacted
and failed and newest and oldest texts. In the newest texts of enacted
bills, there is much more variation in predicted probabilities within
bills.

\begin{figure}
\includegraphics[width=4.1in]{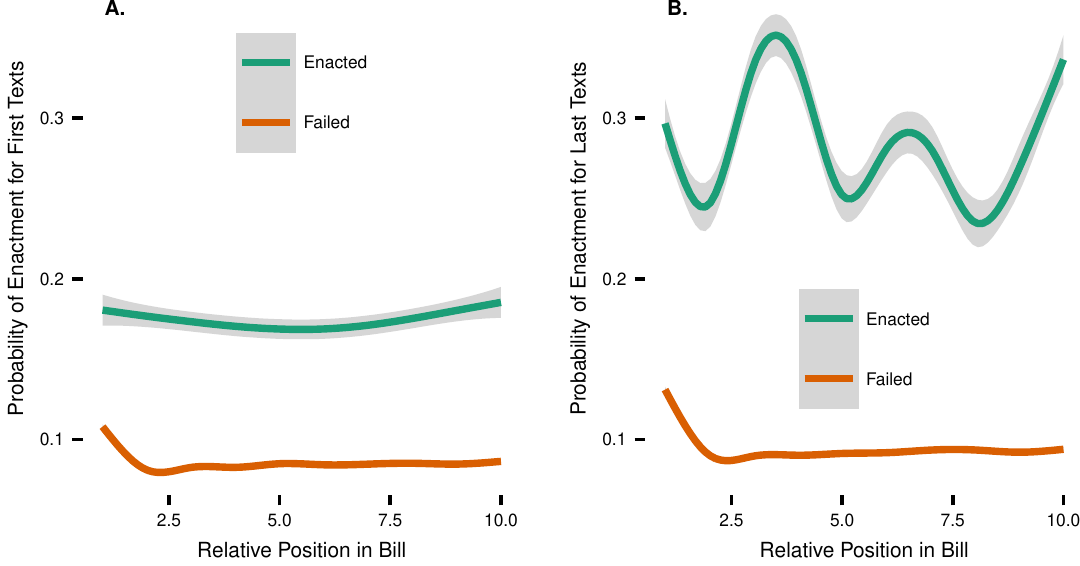}
\caption{Sentence probabilities across bills for oldest data
(\textbf{A.}), newest data (\textbf{B.}). For each bill, we
convert the variable length vectors of predicted sentence probabilities
to \emph{n}-length vectors by sampling \emph{n} evenly-spaced points
from each bill. We set \emph{n=10} because almost every bill is at least
10 sentences long. Then we loess-smooth the resulting points across all
bills to summarize the difference between enacted and failed and newest
and oldest texts.}
\end{figure}

We conducted a partial rank correlation coefficient sensitivity analysis
to estimate the effect of each predictor variable on the predicted
probability of enactment. These are not bivariate correlations between
variables and the predicted probabilities, rather, they are estimates of correlation \emph{after
controlling for} the effect of all other predictor variables, e.g.~the
effect of a bill being introduced in the House is negative after
controlling for the other effects in the model (Fig. 6B.) but bills
introduced in the House are enacted at a 0.043 rate while Senate bills
are enacted at a 0.025 rate. If we stopped with the simple descriptive
statistic we could have incorrectly concluded that introducing a bill in
the House will increase its odds, all else equal.

The two subjects with the largest negative effects are Foreign Trade and
International Finance, and Taxation (Fig. 6A.). Some bills fail because
their content is integrated into other bills and this is especially true
for tax-related bills \cite{yano_textual_2012}. With the oldest data model, increasing bill
length decreases enactment probability but with the newest data the
opposite relationship holds (Fig. 6B.). We repeated the sensitivity
analysis on the model where no text predictions are included
(\emph{GLM}, see S1 Appendix), and found that, under both time conditions, when
we don't control for the probability of the text by
including our language model predictions (\emph{GLM}), longer texts are more negative
than when we control for the text (\emph{w2vGLM}), and that this
difference is much larger for the newest data. This suggests that the
better we capture the probability of the text and control for its
effects, the better we isolate estimates of non-textual effects.

If the bill sponsor's party is the majority party of their chamber, the
probability of the bill is much higher, especially with the oldest data
where the model relies on this as a key signal of success. Increasing
the number of terms the sponsor has served in Congress also has a
positive effect. The predictive model learned interactions as well: the
number of co-sponsors has a stronger positive effect in the Senate for
the newest data and in the House for the oldest data. If the bill text
scored by the language model is in the second session of the Congress,
for the newest data model, this can serve as a signal that a bill is
being updated and thus it has a higher chance of enactment. For the
oldest data, this means the bill was introduced in the second session,
which is not particularly indicative of success or failure.

\begin{figure}
\includegraphics[width=5.3in]{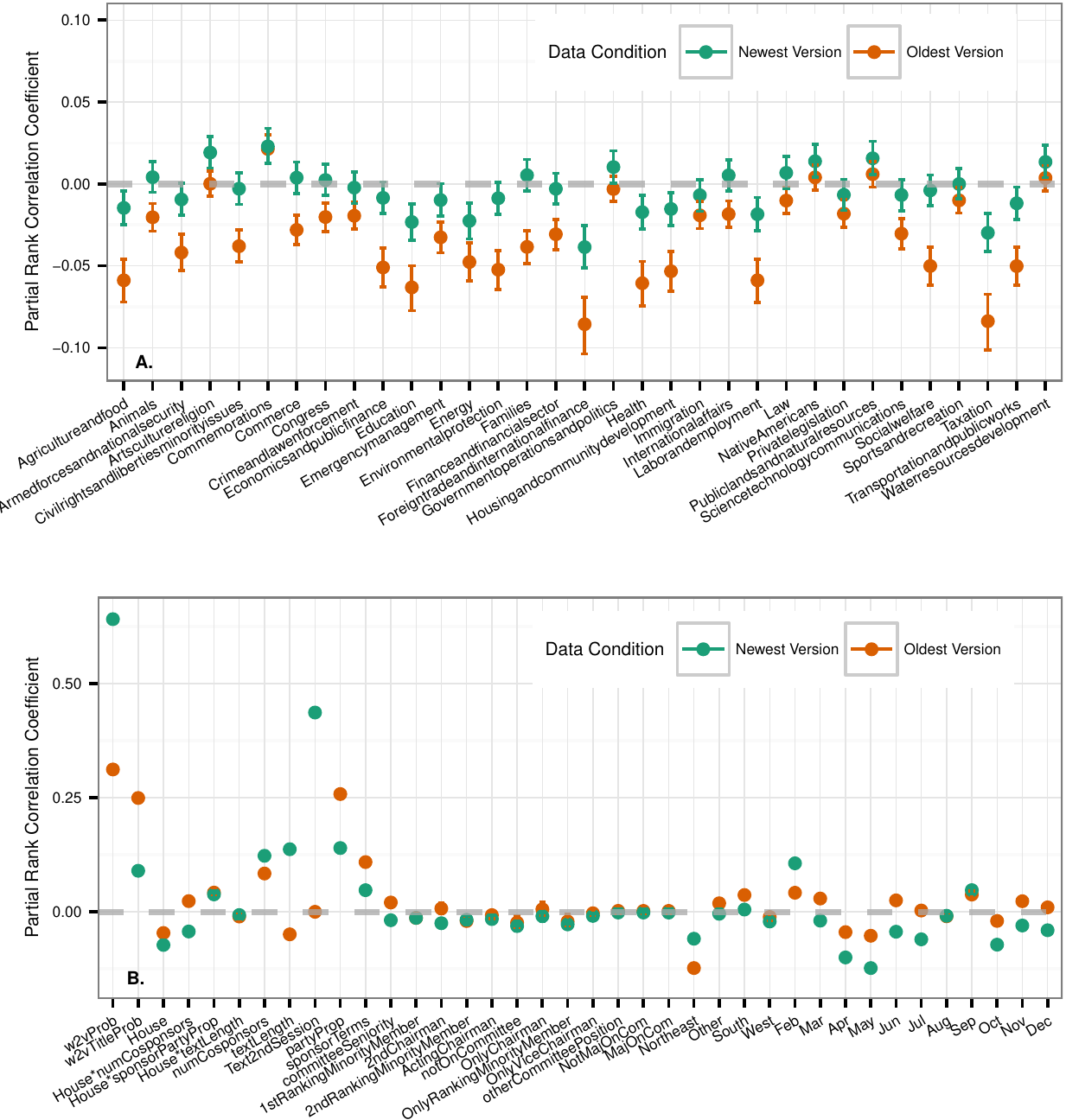}
\caption{Partial rank correlation coefficient estimates between
all \emph{w2vGLM} predictor variables and predicted probabilities (n=55,695, the subset of the observations used to predict the 113th Congress that
had no missing values). Bars represent 95\% confidence intervals.
\textbf{A.} Effects of top subjects. Social Sciences and History
is used as the reference subject so no effect is estimated for that
factor level. \textbf{B.} Effects of all other variables other
than subject. January and North Central are the reference levels for the
month and region factors. This means they are the base level that is not included as a predictor variable itself -- the standard practice when estimating linear models with factor variables. See S1 Appendix for same analysis of \emph{GLM}.}
\end{figure}

\section{Discussion}\label{discussion}

We compared five models across three performance measures and two
data conditions over 14 years.  A model using only bill text outperforms a model
using only bill context for newest data, while context-only outperforms
text-only for oldest data. In all conditions text consistently adds
predictive power.

In addition to accurate predictions, we are able to improve our
understanding of bill content by using a text model designed to explore
differences across chamber and enactment status for important topics.
Our textual analysis serves as an exploratory tool for investigating
subtle distinctions across categories that were previously impossible to
investigate at this scale. The same analysis can be applied to any words
in the large legislative vocabulary. The global sensitivity analysis of
the full model provides insights into the variables affecting predicted
probabilities of enactment. For instance, when predicting bills as they
are first introduced, the text of the bill and the proportion of the
chamber in the bill sponsor's party have similarly strong positive
effects. The full text of the bill is by far the most important
predictor when using the most up-to-date data. The oldest data model
relies more on title predictions than the newest data model, which is understandable given that titles rarely change after bill introduction. Comparing effects across time conditions and across models not including text
suggests that controlling for accurate estimates of the text probability
is important for estimating the effects of non-textual variables.

Although the effect estimates are not causal and estimates on predictors
correlated with each other may be biased, they represent our best
estimates of predictive relationships within a model with the strongest
predictive performance and are thus useful for understanding the process of law-making.
This methodology can be applied to analyze any predictive model by
treating it as a ``black-box'' data-generating process, therefore
predictive power of a model can be optimized and subsequent analysis can
uncover interpretable global relationships between predictors and output. Our
work provides guidance on effectively combining text and context for
prediction \emph{and analysis} of complex systems with highly imbalanced
outcomes that are related to textual data. Our system for determining
the probability of enactment across the thousands of bills
under consideration focuses effort on legislation that is likely to matter, allowing users to identify policy signal amid political and procedural noise.

\section*{Supporting Information}

\paragraph*{S1 Appendix}
\label{S1}
{\bf Supplementary Information.}


\end{document}